# Why Large Language Models and Humans Converge and Diverge in Evaluating Creativity

Pengzhao Lyu[1], Yeun Joon Kim[1,2*], Hanlin Xiao[3,4], Yingyue Luna Luan[5]

[1] Judge Business School, University of Cambridge, Cambridge, CB2 1AG, United Kingdom

[2] Institute of Metabolic Science, School of Clinical Medicine, University of Cambridge, Cambridge, CB2 0QQ, United Kingdom

[3] Department of Computer Science, University of Manchester, Manchester, M13 9PL, United Kingdom

[4] Manchester Institute of Biotechnology, University of Manchester, Manchester, M1 7DN, United Kingdom

[5] School of Business, University of Queensland, Brisbane, QLD 4067, Australia

[*] To whom correspondence may be addressed. Email: yj320@cam.ac.uk

**Abstract**

Despite the growing use of large language models (LLMs) as creativity evaluators, evidence of their alignment with human evaluations remains mixed, raising the question of when and why their judgments converge with or diverge from those of humans. Across three studies and six widely used LLMs, we addressed this gap by identifying the standards underlying LLM creativity evaluation and examining their downstream implications. Study 1 showed that LLMs generally relied on a narrower subset of human creativity evaluation standards. Convergence with human standards was strongest in the novelty dimension, whereas divergence was clearest in the contextual dimension, which captures social, market, and reputational information. Moreover, each LLM exhibited distinct, model-specific standards that varied substantially in breadth. These differences in evaluation standards were reflected in actual creativity judgments. Study 2 ($N$ = 1,103 ideas) showed that LLM evaluations were moderately correlated with human evaluations, and individual LLMs with broader standards better distinguished ideas humans judged as more versus less creative. Study 3 ($N$ = 1,195 participants) showed that LLMs were less sensitive to contextual information: such information significantly altered human creativity ratings but left LLM ratings largely unchanged. Together, our findings help explain the mixed evidence on LLM–human alignment, showing that alignment depends on the evidence a judgment demands and the standards each model applies. Selecting an LLM evaluator is therefore a consequential decision: different models, applying different standards, recognise different ideas as creative.

Large language models (LLMs) are increasingly used not only to generate ideas but also to evaluate ideas across creative and knowledge-work contexts[3–5]. Individuals seek evaluative feedback from LLMs on proposals, designs, and business ideas[5], while organisations use them to screen, compare, and prioritise candidate ideas[3]. In this evaluative role, LLMs are beginning to assume functions previously reserved for humans, influencing which ideas are recognised, developed, and allocated further resources[1,2]. In scientific discovery, for example, AI systems are being developed to assess and rank proposed research ideas, helping determine which ideas move forward to empirical testing and which may be left unexplored[4,14]. Despite the growing use of LLMs as creativity evaluators, we still lack a clear theoretical and empirical understanding of what standards LLMs apply when evaluating creativity and where those standards converge with or diverge from human standards.

Existing research on creativity evaluation has primarily examined the evaluation standards *humans* apply when judging the creativity of ideas, products, and processes. Early work adopted a narrower view, emphasising two core standards for evaluating creativity: novelty and usefulness[1,2,15,16]. From this perspective, creativity evaluation was understood largely as an assessment of the idea itself, focusing on whether an idea was original and whether it served a useful purpose. Later research, however, has shown that human creativity evaluation draws on a broader set of standards[17–20]. Beyond novelty and usefulness, humans also apply standards that capture other qualities of the idea itself, such as hedonic value and technological sophistication, as well as standards that draw on information surrounding the idea, such as social, market, and reputational context[17,18]. Together, this research has identified 26 creativity evaluation standards that capture the diverse forms of evidence humans use when judging creativity (see the Creativity Evaluation Standards column in Table 1).

Building on these established human creativity evaluation standards, our research examines how *LLMs* evaluate creativity. Because LLMs are trained on vast corpora of human-generated data, they might be expected to approximate human creativity judgments. However, emerging research on LLM creativity evaluation has provided mixed evidence[6–13]. Some studies show that LLMs can approximate human creativity judgments in certain settings[6,9,10,12], whereas others find weaker alignment[7,13] or meaningful differences across LLMs[8,11]. These mixed findings suggest that agreement between LLMs and humans in creativity evaluation cannot be understood simply as a question of whether LLMs can or cannot replicate human judgments. Rather, agreement may depend on which evaluation standards are involved. LLMs may, for example, converge with humans on some standards while diverging from them on others, resulting in close alignment in some evaluative contexts but systematic differences in others. Therefore, the first objective of our research is to examine where LLM creativity evaluation standards converge with and diverge from human standards, and whether these patterns are shared across LLMs.

After mapping LLM creativity evaluation standards, a natural next question concerns their implications for downstream creativity judgments. The extent to which LLM evaluation standards converge with or diverge from human standards may determine how closely LLMs track human evaluations and where they systematically deviate from them. When LLMs preserve a broader range of the 26 standards humans use to evaluate creativity, they may better approximate human evaluations by incorporating multiple forms of evidence, including an idea's novelty, usefulness, and surrounding context[15,17,18]. Greater convergence with human standards may therefore enable LLMs to better distinguish outputs that humans judge as more creative from those they judge as less creative. By contrast, divergence from human standards may indicate that LLMs assign different weights to the set of evaluation standards or rely on a narrower subset of those standards when judging creativity. When LLMs systematically underweight standards that humans rely on, they may be less sensitive to information associated with those standards, even when that information substantially shapes human evaluations. The same creative output may therefore be judged differently depending on whether the evaluator incorporates or discounts that information. Therefore, the second objective of our research is to examine how convergence and divergence between LLM and human creativity evaluation standards shape downstream creativity judgments.

We conducted three studies that build on one another, moving from the identification of LLM creativity evaluation standards to the examination of their downstream consequences (Fig. 1). Study 1 addresses the first objective by mapping the creativity evaluation standards of six state-of-the-art LLMs from major AI providers: OpenAI's GPT-4.1 mini[21], xAI's Grok 4 Fast[22], DeepSeek V3[23], Baidu's ERNIE 3.5-8K[24], Anthropic's Claude Haiku 4.5[25], and Google's Gemini 2.5 Flash-Lite[26] (for brevity, we refer to these models hereafter as GPT, Grok, DeepSeek, ERNIE, Claude, and Gemini, respectively). Using an established creativity evaluation standard measure[17], we asked each model to rate the importance of each standard and interpreted those standards rated as important as part of that model's evaluation standard. We first averaged importance ratings across the six models to identify general patterns in LLM creativity evaluation standards, including where they converged with or diverged from human standards. We then analysed each model separately to characterise model-specific variation.

To achieve the second objective, Studies 2 and 3 examine whether and how the convergence and divergence between LLM and human evaluation standards identified in Study 1 are reflected in downstream creativity evaluations. Study 2 investigates the implications of standard convergence in creativity evaluations. We recruited 293 human participants to generate 1,103 creative ideas across three widely used creativity tasks spanning multiple real-world contexts. Following the consensual assessment technique [15], three trained human evaluators and the six LLMs independently evaluated the full set of ideas. By comparing LLM and human creativity ratings, we first examined the alignment between LLMs and humans in creativity evaluation at the aggregate level, and then tested whether LLMs with broader, more human-convergent standards produced creativity ratings that more closely tracked human evaluations. Study 3 ($N$ = 1,195) investigates the implications of standard divergence in creativity evaluations. Using 15 Kickstarter products across multiple product categories, we tested whether information tied to the contextual dimension of creativity evaluation standards—the strongest point of divergence between LLM and human standards identified in Study 1—shifted human creativity evaluations more strongly than LLM evaluations, thereby producing systematic differences in creativity judgments.

## Results

### *LLMs rely on a narrower subset of human creativity evaluation standards*

To address the first objective, Study 1 identified the general pattern of LLM creativity evaluation standards by examining the extent to which LLMs, on average, reproduced human standards for evaluating creativity. Drawing on an established creativity evaluation standard measure[17], we conducted an importance-rating task in which six LLMs rated the importance of 26 standards previously identified as constituting human standards for judging creative ideas (see Table 1 for 26 creativity evaluation standards and their seven corresponding dimensions). Each standard captures a distinct type of information that may serve as evidence of creativity (Supplementary Table 1). Following prior research[17], a standard was considered important—and therefore treated as part of a model's creativity evaluation standard—if the 95% confidence interval around its mean importance rating lay entirely above the scale midpoint. We then averaged importance ratings across models to identify the standards that characterised the general LLM creativity evaluation standard, examine how these standards were distributed across the seven dimensions, and assess their relative importance within the overall ranking.

LLMs exhibited a narrower set of standards for evaluating creativity than humans (Fig. 2). Whereas prior research shows that human creativity evaluation standards encompass all 26 standards[17], the general LLM evaluation standard included only 17, resulting in a 65.38% overlap with the human standards. This result indicates that LLMs did not draw on the same broad range of information that humans use when evaluating creativity. Instead, LLMs preserved only a subset of human creativity evaluation standards, highlighting the need to examine the structure underlying this selectivity.

To characterise the structure, we conceptually categorised the 26 standards into seven dimensions based on the types of information they capture (Table 1). These dimensions distinguish standards that focus on the intrinsic qualities of the idea, such as novelty and usefulness[1,2], from standards that refer to the broader social, market, and reputational context in which the idea is evaluated[18]. This grouping allowed us to examine whether LLMs reproduced human standards broadly across dimensions or instead followed a structured pattern centred on particular dimensions of creativity evaluation.

We found that LLMs converged most strongly with human evaluation standards in the novelty dimension, followed by the hedonic dimension. As shown in Fig. 2, all novelty-related standards were incorporated into the general LLM evaluation standard and dominated the top of the importance ranking. This prioritisation indicates that LLMs substantially preserved the novelty-centred core of human creativity evaluation[1,27,28]. Convergence was not limited to novelty, however, as the hedonic dimension was also consistently represented in the general LLM standard, though these standards received lower importance ratings than novelty-related standards. This suggests that LLMs also drew on information about whether an idea was enjoyable, artistic, or experientially engaging when evaluating creativity. Beyond the novelty and hedonic dimensions, convergence became more limited and selective, with LLMs incorporating only a few standards from the usefulness, concreteness, relational, and technology dimensions.

The clearest divergence between the general LLM standard and human evaluation standards appeared in the contextual dimension. In contrast to humans, LLMs generally excluded contextual standards from their creativity evaluation standard, suggesting that they applied more idea-centred standards for evaluating creativity. As shown in Fig. 2, four of the five contextual standards fell below the importance threshold, with “name brand,” “mass market,” and “fashionability” ranked among the least important standards. “Credibility” was the only contextual standard incorporated into the general LLM standard, with its importance rating just above the threshold. This pattern indicates that LLMs generally treated contextual standards as weak evidence of creativity.

Together, these results suggest that LLMs, in general, apply a narrower, more idea-centred standard for evaluating creativity than humans. Their convergence with human standards was strongest in the novelty dimension and extended to the hedonic dimension, whereas divergence was most evident in the substantial exclusion of contextual standards.

***LLMs exhibit model-specific creativity evaluation standards while preserving a shared evaluative core***

Having established the general pattern of creativity evaluation standards across six LLMs, we next examined each individual model to determine whether LLMs share a universal creativity evaluation standard or instead apply distinct, model-specific standards.

LLMs varied systematically in the standards they applied when evaluating creativity. To examine whether LLMs shared a common creativity evaluation standard or instead relied on distinct model-specific standards, we applied a Cultural Consensus Theory model[29–31] to their importance ratings for the 26 standards. This approach allowed us to identify how many distinct patterns of creativity evaluation standards were present across the six LLMs. As shown in Supplementary Fig. 1, scree plots based on two fit indicators, AIC and BIC, exhibited an elbow at six clusters, with model fit improving substantially from five clusters ($AIC_5$ = 11,844.30, $BIC_5$ = 20,545.41) to six clusters ($AIC_6$ = -1,686.35, $BIC_6$ = 7,706.30). Importantly, the six-cluster classification showed perfect purity, with each cluster containing responses from only one LLM (Supplementary Fig. 2). These results indicate that creativity evaluation standards differed across models, with each LLM exhibiting a distinct model-specific pattern rather than all LLMs sharing a single common standard.

These model-specific creativity evaluation standards differed substantially in their breadth. As shown in Fig. 2, GPT and Grok applied the broadest standards, incorporating 20 standards each and spanning multiple dimensions, such as novelty, usefulness, hedonic, and relational dimensions. DeepSeek and ERNIE exhibited moderate breadth, each incorporating 14 standards and extending beyond the novelty core more selectively, mainly through standards from the hedonic, relational, and technology dimensions. Claude and Gemini applied the narrowest standards, incorporating only 12 and 11 standards, respectively. Their standards remained most tightly anchored to the novelty dimension, with most expansion beyond this core limited to the hedonic dimension. Together, these differences show that LLMs varied considerably in the extent to which they preserved the breadth of human creativity evaluation standards, ranging from relatively broad, multidimensional standards to narrower standards centred more tightly on novelty.

Despite these differences in breadth, the model-specific standards shared several common features. The novelty dimension formed the most stable shared evaluative core, with seven of the eight novelty-related standards ranked among the most important standards across all models. The hedonic dimension was also consistently represented, although it occupied a less central position. At the other end of the ranking, most contextual standards were consistently downweighted across models, indicating a shared evaluative boundary around contextual information. This shared structure suggests that LLMs differed primarily in how broadly each model extended its evaluation standard beyond this shared evaluative core and boundary.

Overall, these results indicate that LLM creativity evaluation standards were organised around a shared evaluative core, encompassing novelty and hedonic standards, while differing in breadth and composition beyond that core. In other words, models shared similar evaluative foundations but varied in how broadly they extended those foundations to incorporate additional standards.

***Downstream consequences of standard convergence***

Study 1 addressed our first objective by examining whether LLMs' reported importance ratings of creativity evaluation standards converged with or diverged from human standards. Building on this foundation, our second objective was to examine whether these patterns of convergence and divergence were reflected in actual creativity evaluations.

Study 2 addressed the implications of standard convergence and proceeded in two parts. In Part 1, we collected 1,103 creative ideas from 293 human participants using three established creativity tasks, including two in-basket exercises that capture workplace problem-solving in realistic business scenarios[32] and one environmental task related to water conservation[33]. Participants generated one idea for each in-basket task and three ideas for the environmental task. In Part 2, we obtained creativity evaluations for the same set of ideas from both human evaluators and LLMs. For the human evaluations, we recruited three trained evaluators and, following the consensual assessment technique[15], averaged their ratings to obtain a final human creativity score for each idea. For the LLM evaluations, each of the six LLMs rated the full set of ideas, yielding model-specific creativity ratings. To test whether the convergence identified in Study 1 was reflected in downstream creativity judgments, we first assessed overall alignment between LLM and human ratings and then examined whether alignment varied across models as a function of the breadth of their creativity evaluation standards.

Consistent with the partial convergence between the general LLM standard and human standards identified in Study 1, LLM evaluations showed moderate alignment with human creativity evaluations. We averaged creativity ratings across the six models to obtain an aggregate LLM creativity rating for each idea, and then correlated these ratings with the human creativity ratings. As shown in Fig. 3, the Pearson correlation between the aggregate LLM ratings and the human creativity ratings was moderate ($r$ = 0.50), indicating that LLM evaluations, in general, partially tracked how humans differentiated more creative ideas from less creative ones, while still leaving room for divergence.

Next, we examined the alignment between each LLM and human creativity evaluations. We correlated each model's creativity ratings with the human creativity ratings across the full set of ideas. Pearson correlation results showed a three-tier pattern (Fig. 3), and Steiger's tests comparing correlation strength across models confirmed that this hierarchy was statistically reliable[34]. GPT and Grok, the two models with the broadest standards identified in Study 1, showed the strongest alignment with human evaluations ($r_{GPT} = 0.52$, $r_{Grok} = 0.49$), exceeding all models with narrower standards in correlation strength (all pairwise comparison $ps \leq 0.042$; Supplementary Table 2), but did not differ significantly from one another ($p = 0.110$). DeepSeek and ERNIE, which showed moderate breadth, occupied the middle tier ($r_{DeepSeek} = 0.46$, $r_{ERNIE} = 0.46$) and aligned more strongly with human evaluations than Claude and Gemini (all pairwise comparison $ps \leq 0.038$; Supplementary Table 2), while not differing significantly from each other ($p = 0.927$). Finally, Claude and Gemini, the two models with the narrowest standards, showed the weakest alignment with human evaluations ($r_{Claude} = 0.41$, $r_{Gemini} = 0.41$) and did not differ significantly from each other ($p = 0.904$).

Together, these results show that convergence between LLM and human creativity evaluation standards carried downstream consequences for actual creativity judgments. At the aggregate level, LLMs showed moderate alignment with human evaluations, consistent with the partial standard convergence identified in Study 1. At the model level, alignment varied systematically with the breadth of LLM creativity evaluation standards, such that models with broader standards better tracked how humans differentiated more creative from less creative ideas.

***Downstream consequences of standard divergence***

We next investigated the downstream consequences of divergence between LLM and human standards. Building on Study 1, we focused on the contextual dimension of creativity evaluation standards, where LLMs diverged most clearly from humans. Whereas humans incorporated contextual standards into their creativity evaluations, LLMs generally assigned less weight to such information, exhibiting more idea-centred evaluation standards. This pattern suggests that information tied to the contextual dimension, such as a product's market positioning, brand prestige, or fashionability, may significantly change human creativity evaluations without similarly altering LLM evaluations. Study 3 ($N$ = 1,195) examined this divergence within product evaluation contexts, where creative ideas are presented as concrete offerings and are naturally accompanied by information related to social, market, and reputational contexts. Specifically, we tested whether contextual information differentially influenced LLM and human creativity evaluations.

Study 3 employed a 2 (Product Description: product-only vs. product-with-context) × 2 (Evaluator Source: human vs. LLM) experimental design. The Product Description factor varied whether evaluators received only a description of the product itself or the same description plus contextual information about the product. Specifically, in the product-only condition, evaluators received a base product description that described the product's features, functions, components, and use. In the product-with-context condition, evaluators received the same base product description plus one additional sentence that provided contextual information about the product. The Evaluator Source factor varied whether the creativity evaluation was performed by human participants or by LLMs.

We created base descriptions for 15 Kickstarter products spanning multiple product categories, using materials provided by each product's creator team. For each product, we also created five one-sentence contextual variants, each corresponding to one of the five contextual standards examined in Study 1 (Supplementary Table 3). Human participants were randomly assigned to evaluate one product in either the product-only condition or the product-with-context condition. In parallel, each LLM evaluated all product descriptions used in the study. For each of the 15 products, the LLMs evaluated one base description and five additional versions of that description, each supplemented with one contextual variant. We then conducted the preregistered multilevel moderation analysis to test whether contextual information affected LLM and human creativity evaluations differently. Creativity ratings were mean-centred within evaluator source, and product was included as a random effect to account for baseline differences in creativity ratings across products.

Results showed that information tied to contextual standards shifted human creativity evaluations but had little effect on LLM evaluations. The interaction between Product Description and Evaluator Source was significant ($b$ = -0.16, $SE$ = 0.04, $p$ < 0.001), indicating that the effect of contextual information differed between human participants and LLMs. As shown in Fig. 4, contextual information significantly increased creativity ratings among human participants ($b$ = 0.18, $SE$ = 0.04, $p$ < 0.001), whereas LLM evaluations remained largely unchanged when contextual information was provided ($b$ = 0.01, $SE$ = 0.02, $p$ = 0.367). This asymmetric response indicates that LLMs were less sensitive to contextual information than humans when evaluating creativity. Building on this finding, we conducted two exploratory multilevel moderation analyses to examine whether this reduced contextual sensitivity varied across LLMs and across types of contextual information.

First, our exploratory analyses showed that reduced sensitivity to contextual information was consistent across LLMs. In multilevel moderation analyses comparing each model with human participants, all interactions were significant, suggesting that contextual information influenced each LLM's creativity evaluations differently than human evaluations (interaction with GPT: $b$ = -0.13, $SE$ = 0.05, $p$ = 0.009; interaction with Grok: $b$ = -0.15, $SE$ = 0.05, $p$ =

0.002; interaction with DeepSeek: $b = -0.18$, $SE = 0.05$, $p < 0.001$; interaction with ERNIE: $b = -0.19$, $SE = 0.05$, $p < 0.001$; interaction with Claude: $b = -0.14$, $SE = 0.05$, $p = 0.004$; interaction with Gemini: $b = -0.20$, $SE = 0.05$, $p < 0.001$). As shown in Fig. 5, this difference reflected a clear asymmetry: contextual information significantly increased creativity ratings among human participants ($b = 0.18$, $SE = 0.04$, $p < 0.001$), whereas none of the LLMs showed a significant change in their evaluations (GPT: $b = 0.05$, $SE = 0.03$, $p = 0.118$; Grok: $b = 0.03$, $SE = 0.03$, $p = 0.411$; DeepSeek: $b = -0.001$, $SE = 0.03$, $p = 0.967$; ERNIE: $b = -0.01$, $SE = 0.03$, $p = 0.805$; Claude: $b = 0.04$, $SE = 0.03$, $p = 0.233$; Gemini: $b = -0.03$, $SE = 0.03$, $p = 0.410$).

Second, reduced sensitivity to contextual information was most evident for the contextual standards that LLMs discounted most strongly in Study 1. When each contextual variant was compared with the product-only condition, contextual sensitivity differed significantly between LLMs and human participants for the “name brand” ($b = -0.29$, $SE = 0.06$, $p < 0.001$), “mass market” ($b = -0.19$, $SE = 0.07$, $p = 0.004$), and “fashionability” ($b = -0.33$, $SE = 0.07$, $p < 0.001$) variants. As shown in Fig. 6, contextual information significantly increased human creativity evaluations for these variants (name brand: $b = 0.29$, $SE = 0.06$, $p < 0.001$; mass market: $b = 0.16$, $SE = 0.06$, $p = 0.009$; fashionability: $b = 0.33$, $SE = 0.06$, $p < 0.001$), whereas LLM evaluations remained largely unchanged (name brand: $b = 0.01$, $SE = 0.02$, $p = 0.720$; mass market: $b = -0.03$, $SE = 0.02$, $p = 0.253$; fashionability: $b = 0.01$, $SE = 0.02$, $p = 0.720$). By contrast, we did not observe significant differences in contextual sensitivity for the “social approval” ($b = -0.001$, $SE = 0.07$, $p = 0.983$) or “credibility” ($b = -0.01$, $SE = 0.07$, $p = 0.937$) variants. These results indicate that divergence in contextual sensitivity was not uniform across all forms of contextual information, but was clearest for information tied to brand prestige, market positioning, and fashionability.

Overall, Study 3 showed that divergence between LLM and human creativity evaluation standards was reflected in systematic downstream differences in creativity judgments. Consistent with the more idea-centred standards identified in Study 1, contextual information shifted human creativity evaluations but left LLM evaluations largely unchanged, indicating that LLMs were less sensitive to contextual information that humans used when evaluating creativity.

## Discussion

Despite the growing use of LLMs as creativity evaluators, it remains unclear how they judge creativity and when and why their judgments align with or differ from those of humans. Across three studies, we addressed these questions by examining the standards that guide LLM creativity evaluation and the downstream consequences of their convergence with and divergence from human standards. Study 1 showed that LLMs generally applied narrower, more idea-centred standards than humans. Whereas human creativity evaluation encompasses both intrinsic qualities of the idea and information surrounding it, including social, market, and reputational factors, LLMs placed substantially less weight on these contextual standards. We also found meaningful variation across LLMs in the breadth of their evaluation standards. Some models, such as GPT and Grok, incorporated a broader set of standards that more closely approximated human standards, whereas others relied on narrower standards that centred more strongly on the idea itself.

These differences in evaluation standards were reflected in downstream creativity judgments. Study 2 showed that LLM evaluations were moderately correlated with human evaluations, suggesting both a shared reliance on idea-centred standards and meaningful divergence arising partly because LLMs placed less weight on contextual standards that humans incorporated into their judgments. Consistent with this interpretation, models with broader, more human-convergent standards exhibited stronger alignment with human evaluations. Study 3 provided further evidence by showing that information tied to contextual standards shifted human creativity evaluations but left LLM evaluations largely unchanged. Together, these findings suggest that LLMs may approximate human creativity evaluations when a task relies on standards they share with humans, but are more likely to diverge when the task requires contextual information that they tend to underweight.

This work provides important and timely contributions to both theory and practice. First, it advances emerging research on LLMs and creativity evaluation by helping explain the inconsistent findings on LLM–human alignment. Existing studies have largely focused on whether LLMs can replicate human creativity judgments, yet findings have varied substantially[6–13]. Our findings suggest that this variation may arise from differences in the creativity evaluation standards applied by LLMs and humans. More specifically, LLMs and humans differ not only in the breadth of the standards they apply, but also in the kinds of evidence they treat as informative for creativity. Humans draw on both the intrinsic qualities of an idea and the social, market, and reputational context in which it is situated, whereas LLMs are more strongly anchored to the idea itself and assign less weight to contextual information. As a result, LLMs may more closely approximate human evaluations when a task primarily focuses on the intrinsic qualities of an idea, such as its novelty, unexpectedness, or conceptual promise, but may diverge when creativity judgments require attention to contextual evidence, such as whether the idea appears credible to stakeholders, aligns with market expectations, or receives social approval. By identifying the standards that underlie LLM evaluations, this work moves beyond the question of *whether* LLMs can reproduce human creativity judgments and advances our understanding of *when* and *why* their evaluations converge with or diverge from human judgments.

Second, our work reveals a discrepancy between formal theories of creativity, which equally emphasise both novelty and usefulness, and LLM evaluation standards, which place greater emphasis on novelty. Creativity research has long emphasised both novelty and usefulness as defining features of creative ideas[15,27,28,35]. Yet, across the six LLMs we examined, novelty-related standards formed the most central and stable component of creativity evaluation, whereas usefulness-related standards were incorporated more selectively and occupied a more peripheral position. This asymmetry suggests that LLMs may not simply reproduce formal theories of creativity. Instead, they may assign greater weight to standards that are more explicitly and consistently associated with creativity in the textual data on which they are trained. Novelty may be especially salient because creativity is frequently described using terms such as original, unconventional, breakthrough, or paradigm-shifting. Other standards may be less visible because they are less consistently framed as direct evidence of creativity or are applied more implicitly in human judgments. For example, feasibility and functionality may signal creative value in some contexts but ordinary practicality or incremental improvement in others. Similarly, factors such as social approval and brand prestige may shape creativity judgments without being explicitly identified as reasons an idea is creative. Human-generated training data may therefore provide a partial and selective representation of human creativity evaluation, preserving standards that are readily verbalised and repeatedly associated with creativity while attenuating those that are applied more tacitly, situationally, or institutionally. Accordingly, LLM creativity evaluation should not be treated as a neutral reflection of either formal theories of creativity or human evaluative practice. Rather, it may disproportionately reflect the standards most visible in textual discourse while underrepresenting those that humans use in practice but do not consistently articulate as evidence of creativity.

Third, our work contributes to ongoing discussions on leveraging LLMs as creativity evaluators in real-world contexts. As organisations and individuals increasingly rely on LLMs for creativity evaluation[3,5], our findings suggest that their effectiveness depends on the fit between the standards a model emphasises and those relevant to a given evaluation. Because LLMs generally applied more idea-centred standards and were less sensitive to contextual information, they may be especially useful when the evaluation should focus on the idea itself while minimising the influence of surrounding contextual signals. In early-stage scientific proposal screening, for example, LLM evaluations may help assess the originality and intellectual merit of a proposal without giving undue weight to the proposer's status, institutional affiliation, or disciplinary prominence[36]. In other settings, however, contextual information may appropriately form part of how creativity is evaluated rather than being merely peripheral[18]. For example, when evaluating alternative product ideas, organisations may reasonably consider whether those ideas align with social norms, appear credible to stakeholders, or have the potential for broad market adoption. In such cases, LLM evaluations may be incomplete or even misleading unless their judgments are complemented by human evaluations. Thus, LLM creativity evaluation should not be treated as universally applicable. Its effectiveness depends on whether the standards emphasised by the model align with the standards that should guide the evaluation at hand.

Lastly, our findings caution against treating different LLMs as interchangeable creativity evaluators. Although discussions of LLM creativity evaluation often treat "LLMs" as a single technology category, our results show that different models apply substantially different standards when judging creativity. As a result, choosing one model over another may meaningfully shape which ideas are recognised as creative. For researchers, this suggests that future work should more carefully distinguish between patterns shared across LLMs and those specific to individual models. For organisations and individuals, it suggests that the relevant question is not merely whether to implement an LLM evaluator, but which model's evaluation standards best align with the type of creativity they aim to recognise.

Despite these contributions, this research has limitations that can stimulate future work. First, consistent with prior research on LLM behaviour[37,38], Study 1 assessed LLM creativity evaluation standards by examining model responses to established psychological measures. This approach allowed us to systematically identify which standards LLMs reported as important and where those standards converged with or diverged from human standards. However, prior research suggests that, as with humans, the standards LLMs report as important may not always correspond to the standards that guide their behaviour across contexts[39], such as their actual creativity evaluations. Our subsequent studies addressed this issue in part by following a progressive design in which Study 1 identified LLM evaluation standards and Studies 2 and 3 examined their downstream consequences. Future research could build on this approach by measuring or manipulating specific evaluation standards within a single study to examine more directly how particular standards influence LLM creativity judgments.

Moreover, by examining six representative state-of-the-art LLMs from major providers, our research offers initial evidence on how leading LLMs evaluate creativity. At the same time, the broader LLM ecosystem includes many other models that may differ in architecture, training data, fine-tuning procedures, alignment methods, or deployment settings. Future research could examine whether the patterns identified here extend to open-source, domain-specialised, and multimodal models, as well as models configured for specific organisational or creative tasks.

Finally, while our work identifies how LLMs evaluate creativity, it leaves open the broader question of how their growing role as creativity evaluators may influence humans. In practice, LLMs often operate interactively, providing feedback, explanations, and suggestions that people may incorporate into their evaluations. This raises an important question that extends beyond whether LLMs align with humans: whether repeated interaction with LLM evaluators changes the standards humans use when judging creativity. Given the persuasive power of LLMs[40–43], repeated exposure to their evaluations may shift how people evaluate creativity over time, potentially creating a self-reinforcing loop in which LLM standards shape human standards and those human responses are later reflected in future model training data. Although we did not examine this possibility directly, our findings provide a foundation for studying it by identifying the creativity evaluation standards that LLMs currently emphasise and underweight. Future research could therefore investigate whether exposure to LLM evaluations shifts human standards, which standards are most susceptible to such influence, and whether this influence amplifies or reduces divergence between LLM and human creativity evaluation over time.

## References

1. Zhou, J., Wang, X. M., Bavato, D., Tasselli, S. & Wu, J. Understanding the Receiving Side of Creativity: A Multidisciplinary Review and Implications for Management Research. *J. Manag.* **45**, 2570–2595 (2019).

2. Zhou, J., Wang, X. M., Song, L. J. & Wu, J. Is it new? Personal and contextual influences on perceptions of novelty and creativity. *J. Appl. Psychol.* **102**, 180–202 (2017).

3. Chen, Z. & Chan, J. Large Language Model in Creative Work: The Role of Collaboration Modality and User Expertise. *Manag. Sci.* **70**, 9101–9117 (2024).

4. Lu, C. *et al.* Towards end-to-end automation of AI research. *Nature* **651**, 914–919 (2026).

5. Luan, Y. L., Kim, Y. J. & Zhou, J. Augmented Learning for Joint Creativity in Human-GenAI Co-Creation. *Inf. Syst. Res.* https://doi.org/10.1287/isre.2024.0984 (2025).

6. DiStefano, P. V., Patterson, J. D. & Beaty, R. E. Automatic Scoring of Metaphor Creativity with Large Language Models. *Creat. Res. J.* **37**, 555–569 (2025).

7. Kranzle, T. & Sharratt, K. Evaluating Creative Output With Generative Artificial Intelligence: Comparing GPT Models and Human Experts in Idea Evaluation. *Creat. Innov. Manag.* **34**, 991–1012 (2025).

8. Li, R., Zhu, C., Xu, B., Wang, X. & Mao, Z. Automated Creativity Evaluation for Large Language Models: A Reference-Based Approach. in *Findings of the Association for Computational Linguistics: EMNLP 2025* (eds Christodoulopoulos, C., Chakraborty, T., Rose, C. & Peng, V.) 21475–21488 (Association for Computational Linguistics, Suzhou, China, 2025). doi:10.18653/v1/2025.findings-emnlp.1171.

9. Organisciak, P., Acar, S., Dumas, D. & Berthiaume, K. Beyond semantic distance: Automated scoring of divergent thinking greatly improves with large language models. *Think. Ski. Creat.* **49**, 101356 (2023).

10. Orwig, W., Edenbaum, E. R., Greene, J. D. & Schacter, D. L. The Language of Creativity: Evidence from Humans and Large Language Models. *J. Creat. Behav.* **58**, 128–136 (2024).

11. Saretzki, J. & Benedek, M. Investigating the Validity Evidence of Automated Scoring Methods for Divergent Thinking Assessments. *Creat. Res. J.* **0**, 1–17 (2026).

12. Goecke, B. *et al.* Automated Scoring of Scientific Creativity in German. *J. Creat. Behav.* **58**, 321–327 (2024).

13. Chakrabarty, T., Laban, P., Agarwal, D., Muresan, S. & Wu, C.-S. Art or Artifice? Large Language Models and the False Promise of Creativity. in *Proceedings of the 2024 CHI Conference on Human Factors in Computing Systems* 1–34 (Association for Computing Machinery, New York, NY, USA, 2024). doi:10.1145/3613904.3642731.

14. Gottweis, J. *et al.* Accelerating scientific discovery with Co-Scientist. *Nature* 1–3 (2026) doi:10.1038/s41586-026-10644-y.

15. Amabile, T. M. Social psychology of creativity: A consensual assessment technique. *J. Pers. Soc. Psychol.* **43**, 997–1013 (1982).

16. George, J. M. Creativity in Organizations. *Acad. Manag. Ann.* **1**, 439–477 (2007).

17. Loewenstein, J. & Mueller, J. Implicit Theories of Creative Ideas: How Culture Guides Creativity Assessments. *Acad. Manag. Discov.* **2**, 320–348 (2016).

18. Mueller, J., Melwani, S., Loewenstein, J. & Deal, J. J. Reframing the Decision-Makers' Dilemma: Towards a Social Context Model of Creative Idea Recognition. *Acad. Manage. J.* **61**, 94–110 (2018).

19. Sternberg, R. J. Implicit theories of intelligence, creativity, and wisdom. *J. Pers. Soc. Psychol.* **49**, 607–627 (1985).
20. Elsbach, K. D. & Kramer, R. M. Assessing Creativity in Hollywood Pitch Meetings: Evidence for a Dual-Process Model of Creativity Judgments. *Acad. Manage. J.* **46**, 283–301 (2003).
21. OpenAI. GPT-4.1 mini. (2025).
22. xAI. Grok 4 Fast. (2025).
23. DeepSeek. DeepSeek V3. (2025).
24. Baidu. ERNIE 3.5-8k. (2025).
25. Anthropic. Claude Haiku 4.5. (2025).
26. Google. Gemini 2.5 Flash-Lite. (2025).
27. Anderson, N., Potočnik, K. & Zhou, J. Innovation and Creativity in Organizations: A State-of-the-Science Review, Prospective Commentary, and Guiding Framework. *J. Manag.* **40**, 1297–1333 (2014).
28. Harvey, S. & Berry, J. W. Toward a Meta-Theory of Creativity Forms: How Novelty and Usefulness Shape Creativity. *Acad. Manage. Rev.* **48**, 504–529 (2023).
29. Anders, R., Oravecz, Z. & Batchelder, W. H. Cultural consensus theory for continuous responses: A latent appraisal model for information pooling. *J. Math. Psychol.* **61**, 1–13 (2014).
30. France, S. L., Sharif Vaghefi, M. & Batchelder, W. H. FlexCCT: A Methodological Framework and Software for Ratings Analysis and Wisdom of the Crowd Applications. *IEEE Trans. Comput. Soc. Syst.* **5**, 358–370 (2018).

31. Romney, A. K., Weller, S. C. & Batchelder, W. H. Culture as Consensus: A Theory of Culture and Informant Accuracy. *Am. Anthropol.* **88**, 313–338 (1986).

32. Shalley, C. E. Effects of productivity goals, creativity goals, and personal discretion on individual creativity. *J. Appl. Psychol.* **76**, 179–185 (1991).

33. OECD. *PISA 2022 Assessment and Analytical Framework.* https://www.oecd.org/en/publications/pisa-2022-assessment-and-analytical-framework_dfe0bf9c-en.html (2023) doi:10.1787/dfe0bf9c-en.

34. Steiger, J. H. Tests for comparing elements of a correlation matrix. *Psychol. Bull.* **87**, 245–251 (1980).

35. Zhou, J. & Hoever, I. J. Research on Workplace Creativity: A Review and Redirection. *Annu. Rev. Organ. Psychol. Organ. Behav.* **1**, 333–359 (2014).

36. Bol, T., de Vaan, M. & van de Rijt, A. The Matthew effect in science funding. *Proc. Natl. Acad. Sci.* **115**, 4887–4890 (2018).

37. Lu, J. G., Song, L. L. & Zhang, L. D. Cultural tendencies in generative AI. *Nat. Hum. Behav.* 1–10 (2025) doi:10.1038/s41562-025-02242-1.

38. Mei, Q., Xie, Y., Yuan, W. & Jackson, M. O. A Turing test of whether AI chatbots are behaviorally similar to humans. *Proc. Natl. Acad. Sci.* **121**, e2313925121 (2024).

39. Wang, X., Zhou, Y. & Zhou, G. Large language models exhibit stigmatizing behaviour in contextual judgements of health conditions. *Nat. Health* 1–13 (2026) doi:10.1038/s44360-026-00164-4.

40. Hackenburg, K. *et al.* The levers of political persuasion with conversational artificial intelligence. *Science* **390**, eaea3884 (2025).

41. Lin, H. *et al.* Persuading voters using human–artificial intelligence dialogues. *Nature* **648**, 394–401 (2025).

42. Lyu, P., Kim, Y. J., Luan, Y. L. & Choi, J. Morally Programmed LLMs Reshape Human Morality. Preprint at https://doi.org/10.48550/arXiv.2604.10222 (2026).

43. Salvi, F., Horta Ribeiro, M., Gallotti, R. & West, R. On the conversational persuasiveness of GPT-4. *Nat. Hum. Behav.* 1–9 (2025) doi:10.1038/s41562-025-02194-6.

**Figure 1. LLM creativity evaluation standards and their downstream consequences for creativity judgments.** Panel (A) presents Study 1, in which six state-of-the-art LLMs rated the importance of 26 human creativity evaluation standards spanning seven dimensions. The general LLM standard incorporated 17 standards, showing 65.38% overlap with human standards. LLMs converged most strongly with human standards in the novelty dimension and diverged most clearly in the contextual dimension. The breadth of standards varied substantially across LLMs. Panel (B) reports Study 2, in which 293 participants generated 1,103 ideas that were independently evaluated by three trained human evaluators and the six LLMs. Aggregate LLM ratings were moderately correlated with human ratings, and models with broader, more human-convergent standards showed stronger alignment with human evaluations. Panel (C) summarises Study 3, in which 1,195 human participants and the six LLMs evaluated 15 Kickstarter products presented either with information about the product's intrinsic qualities, such as functionality, or with the same information supplemented by contextual information. Contextual information increased human creativity ratings but did not reliably change LLM ratings, resulting in significantly lower contextual sensitivity among LLMs than among human participants.

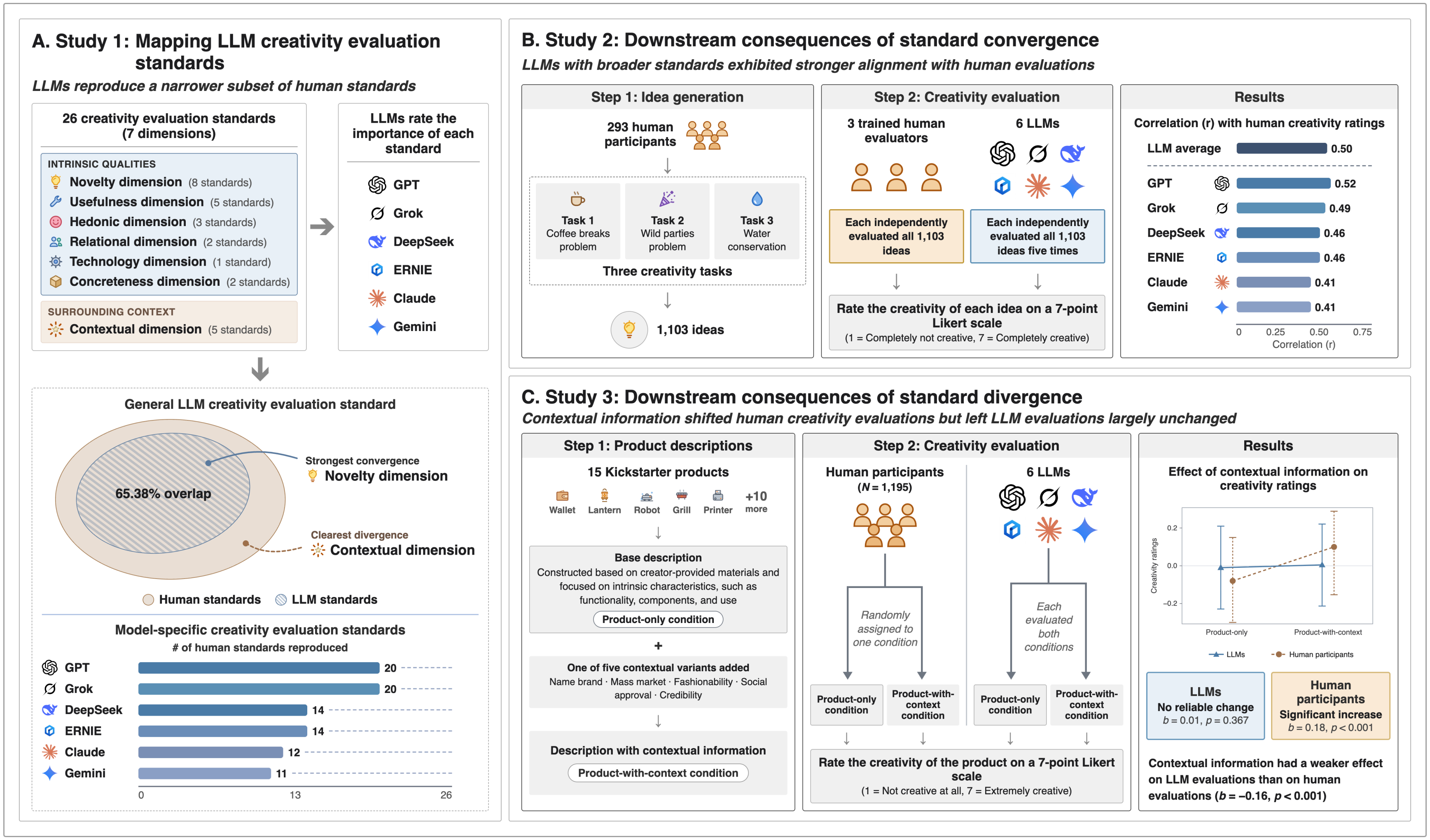

A. Study 1: Mapping LLM creativity evaluation standards
LLMs reproduce a narrower subset of human standards
26 creativity evaluation standards (7 dimensions)
INTRINSIC QUALITIES
Novelty dimension (8 standards)
Usefulness dimension (5 standards)
Hedonic dimension (3 standards)
Relational dimension (2 standards)
Technology dimension (1 standard)
Concreteness dimension (2 standards)
SURROUNDING CONTEXT
Contextual dimension (5 standards)
LLMs rate the importance of each standard
GPT
Grok
DeepSeek
ERNIE
Claude
Gemini
General LLM creativity evaluation standard
65.38% overlap
Strongest convergence
Novelty dimension
Clearest divergence
Contextual dimension
Human standards
LLM standards
Model-specific creativity evaluation standards
# of human standards reproduced
GPT 20
Grok 20
DeepSeek 14
ERNIE 14
Claude 12
Gemini 11
0
13
26
B. Study 2: Downstream consequences of standard convergence
LLMs with broader standards exhibited stronger alignment with human evaluations
Step 1: Idea generation
293 human participants
Task 1 Coffee breaks problem
Task 2 Wild parties problem
Task 3 Water conservation
Three creativity tasks
1,103 ideas
Step 2: Creativity evaluation
3 trained human evaluators
6 LLMs
Each independently evaluated all 1,103 ideas
Each independently evaluated all 1,103 ideas five times
Rate the creativity of each idea on a 7-point Likert scale
(1 = Completely not creative, 7 = Completely creative)
Results
Correlation (r) with human creativity ratings
LLM average 0.50
GPT 0.52
Grok 0.49
DeepSeek 0.46
ERNIE 0.46
Claude 0.41
Gemini 0.41
0
0.25
0.50
0.75
Correlation (r)
C. Study 3: Downstream consequences of standard divergence
Contextual information shifted human creativity evaluations but left LLM evaluations largely unchanged
Step 1: Product descriptions
15 Kickstarter products
Wallet
Lantern
Robot
Grill
Printer
+10 more
Base description
Constructed based on creator-provided materials and focused on intrinsic characteristics, such as functionality, components, and use
Product-only condition
+
One of five contextual variants added
Name brand · Mass market · Fashionability · Social approval · Credibility
Description with contextual information
Product-with-context condition
Step 2: Creativity evaluation
Human participants
(N = 1,195)
6 LLMs
Randomly assigned to one condition
Each evaluated both conditions
Product-only condition
Product-with-context condition
Product-only condition
Product-with-context condition
Rate the creativity of the product on a 7-point Likert scale
(1 = Not creative at all, 7 = Extremely creative)
Results
Effect of contextual information on creativity ratings
Creativity ratings
0.2
0.0
−0.2
Product-only
Product-with-context
LLMs
Human participants
LLMs
No reliable change
b = 0.01, p = 0.367
Human participants
Significant increase
b = 0.18, p < 0.001
Contextual information had a weaker effect on LLM evaluations than on human evaluations (b = −0.16, p < 0.001)

**Figure 2. LLM creativity evaluation standards.** Panel (A) shows the average importance ratings across the six LLMs for the 26 creativity evaluation standards, organised by dimension. Within each dimension, the “Average” row reports the mean importance rating across all standards in that dimension. Each cell reports the mean importance rating with the 95% confidence interval in parentheses. Panel (B) shows the corresponding rankings separately for each individual LLM. Standards within each column are ordered by mean importance rating, from highest to lowest. The LLM average column reports the mean importance rating across all six LLMs. Thick borders separate standards whose confidence intervals were entirely above the midpoint from the remaining standards within each column. In both panels, cell colours indicate mean importance ratings on the 7-point Likert scale, with warmer colours indicating higher ratings and cooler colours indicating lower ratings. Symbols indicate the position of the 95% confidence interval relative to the scale midpoint: ▲ = entirely above the midpoint, ○ = includes the midpoint, and ▼ = entirely below the midpoint.

## A Importance of LLM creativity evaluation standards by dimension

| Cue Dimension | Cue | LLM average | GPT | Grok | DeepSeek | ERNIE | Claude | Gemini |
|---|---|---|---|---|---|---|---|---|
| Novelty dimension | ***Average*** | 6.07 [5.56, 6.58] ▲ | 6.54 [6.53, 6.55] ▲ | 6.68 [6.68, 6.69] ▲ | 5.46 [5.46, 5.46] ▲ | 5.89 [5.87, 5.91] ▲ | 5.67 [5.67, 5.67] ▲ | 6.17 [6.17, 6.17] ▲ |
| | Breakthrough | 6.59 [6.03, 7.15] ▲ | 7.00 [7.00, 7.00] ▲ | 7.00 [7.00, 7.00] ▲ | 6.33 [6.33, 6.33] ▲ | 6.57 [6.54, 6.60] ▲ | 5.67 [5.67, 5.67] ▲ | 7.00 [7.00, 7.00] ▲ |
| | Combination | 6.00 [5.38, 6.63] ▲ | 6.36 [6.34, 6.37] ▲ | 6.66 [6.66, 6.67] ▲ | 5.00 [5.00, 5.00] ▲ | 6.00 [6.00, 6.00] ▲ | 5.67 [5.67, 5.67] ▲ | 6.33 [6.33, 6.33] ▲ |
| | Paradigm shift | 6.77 [6.39, 7.15] ▲ | 7.00 [7.00, 7.00] ▲ | 7.00 [7.00, 7.00] ▲ | 7.00 [7.00, 7.00] ▲ | 6.28 [6.25, 6.30] ▲ | 6.33 [6.33, 6.33] ▲ | 7.00 [7.00, 7.00] ▲ |
| | Rarity | 6.19 [5.41, 6.98] ▲ | 6.68 [6.67, 6.69] ▲ | 7.00 [7.00, 7.00] ▲ | 5.00 [5.00, 5.00] ▲ | 6.15 [6.11, 6.18] ▲ | 5.67 [5.67, 5.67] ▲ | 6.67 [6.67, 6.67] ▲ |
| | Repurposing | 6.22 [5.70, 6.75] ▲ | 6.67 [6.67, 6.67] ▲ | 7.00 [7.00, 7.00] ▲ | 5.67 [5.67, 5.67] ▲ | 6.00 [6.00, 6.00] ▲ | 6.00 [6.00, 6.00] ▲ | 6.00 [6.00, 6.00] ▲ |
| | Updates tradition | 5.54 [4.93, 6.16] ▲ | 6.33 [6.33, 6.33] ▲ | 6.00 [6.00, 6.00] ▲ | 4.67 [4.67, 4.67] ▲ | 5.26 [5.23, 5.30] ▲ | 5.33 [5.33, 5.33] ▲ | 5.67 [5.67, 5.67] ▲ |
| | Potential | 6.62 [6.12, 7.12] ▲ | 7.00 [7.00, 7.00] ▲ | 7.00 [7.00, 7.00] ▲ | 6.67 [6.67, 6.67] ▲ | 6.07 [6.04, 6.09] ▲ | 6.00 [6.00, 6.00] ▲ | 7.00 [7.00, 7.00] ▲ |
| | Surprise | 4.59 [3.61, 5.58] ○ | 5.27 [5.18, 5.37] ▲ | 5.80 [5.77, 5.84] ▲ | 3.33 [3.33, 3.33] ▼ | 4.82 [4.70, 4.94] ▲ | 4.67 [4.67, 4.67] ▲ | 3.67 [3.67, 3.67] ▼ |
| Usefulness dimension | ***Average*** | 4.13 [3.54, 4.71] ○ | 4.85 [4.83, 4.86] ▲ | 4.74 [4.72, 4.75] ▲ | 4.20 [4.20, 4.20] ▲ | 3.66 [3.64, 3.67] ▼ | 3.59 [3.58, 3.60] ▼ | 3.73 [3.73, 3.73] ▼ |
| | Ease of use | 4.03 [3.42, 4.65] ○ | 4.98 [4.96, 5.01] ▲ | 4.29 [4.25, 4.33] ▲ | 4.00 [4.00, 4.00] ○ | 4.00 [3.96, 4.04] ○ | 3.25 [3.22, 3.28] ▼ | 3.67 [3.67, 3.67] ▼ |
| | Feasibility | 3.35 [2.71, 3.98] ▼ | 3.78 [3.72, 3.83] ▼ | 3.42 [3.37, 3.48] ▼ | 3.67 [3.67, 3.67] ▼ | 2.54 [2.51, 2.58] ▼ | 2.67 [2.67, 2.67] ▼ | 4.00 [4.00, 4.00] ○ |
| | Functionality | 4.55 [3.56, 5.53] ○ | 5.67 [5.67, 5.67] ▲ | 5.67 [5.67, 5.67] ▲ | 4.67 [4.67, 4.67] ▲ | 3.95 [3.90, 4.00] ▼ | 3.67 [3.67, 3.67] ▼ | 3.67 [3.67, 3.67] ▼ |
| | Variety | 5.03 [4.57, 5.49] ▲ | 5.51 [5.47, 5.54] ▲ | 5.64 [5.62, 5.66] ▲ | 4.67 [4.67, 4.67] ▲ | 5.00 [5.00, 5.01] ▲ | 4.69 [4.67, 4.72] ▲ | 4.67 [4.67, 4.67] ▲ |
| | Wide use | 3.68 [2.83, 4.53] ○ | 4.30 [4.28, 4.32] ▲ | 4.67 [4.66, 4.68] ▲ | 4.00 [4.00, 4.00] ○ | 2.78 [2.75, 2.81] ▼ | 3.67 [3.67, 3.67] ▼ | 2.67 [2.67, 2.67] ▼ |
| Hedonic dimension | ***Average*** | 5.22 [4.64, 5.80] ▲ | 5.91 [5.89, 5.92] ▲ | 5.76 [5.74, 5.78] ▲ | 4.67 [4.66, 4.67] ▲ | 5.32 [5.31, 5.33] ▲ | 5.11 [5.11, 5.11] ▲ | 4.56 [4.56, 4.56] ▲ |
| | Artistry | 5.42 [4.52, 6.33] ▲ | 6.27 [6.25, 6.30] ▲ | 6.27 [6.21, 6.34] ▲ | 5.00 [4.99, 5.00] ▲ | 5.66 [5.66, 5.67] ▲ | 5.33 [5.33, 5.33] ▲ | 4.00 [4.00, 4.00] ○ |
| | Joy | 4.84 [4.16, 5.53] ▲ | 5.77 [5.74, 5.80] ▲ | 5.33 [5.33, 5.33] ▲ | 4.00 [4.00, 4.00] ○ | 4.62 [4.59, 4.65] ▲ | 5.00 [5.00, 5.00] ▲ | 4.33 [4.33, 4.33] ▲ |
| | Experience | 5.39 [5.04, 5.74] ▲ | 5.68 [5.67, 5.69] ▲ | 5.67 [5.66, 5.68] ▲ | 5.00 [5.00, 5.00] ▲ | 5.67 [5.65, 5.69] ▲ | 5.00 [5.00, 5.00] ▲ | 5.33 [5.33, 5.33] ▲ |
| Relational dimension | ***Average*** | 3.96 [3.31, 4.61] ○ | 4.92 [4.90, 4.93] ▲ | 4.33 [4.33, 4.33] ▲ | 4.00 [4.00, 4.00] ○ | 3.84 [3.83, 3.86] ▼ | 3.17 [3.17, 3.17] ▼ | 3.50 [3.50, 3.50] ▼ |
| | Social interaction | 4.55 [3.98, 5.12] ○ | 5.33 [5.33, 5.33] ▲ | 4.67 [4.67, 4.67] ▲ | 4.33 [4.33, 4.33] ▲ | 4.65 [4.63, 4.66] ▲ | 3.67 [3.67, 3.67] ▼ | 4.67 [4.67, 4.67] ▲ |
| | Harmony | 3.37 [2.50, 4.24] ○ | 4.50 [4.46, 4.53] ▲ | 4.00 [4.00, 4.00] ○ | 3.67 [3.67, 3.67] ▼ | 3.04 [3.01, 3.06] ▼ | 2.67 [2.67, 2.67] ▼ | 2.33 [2.33, 2.33] ▼ |
| Technology dimension | High tech | 4.11 [3.29, 4.92] ○ | 4.98 [4.96, 5.00] ▲ | 4.33 [4.33, 4.33] ▲ | 4.67 [4.67, 4.67] ▲ | 4.33 [4.32, 4.34] ▲ | 3.33 [3.33, 3.33] ▼ | 3.00 [3.00, 3.00] ▼ |
| Concreteness dimension | ***Average*** | 3.13 [2.76, 3.50] ▼ | 3.28 [3.26, 3.30] ▼ | 2.83 [2.83, 2.83] ▼ | 3.67 [3.67, 3.67] ▼ | 3.28 [3.26, 3.31] ▼ | 2.70 [2.69, 2.71] ▼ | 3.00 [3.00, 3.00] ▼ |
| | Observability | 2.54 [1.68, 3.40] ▼ | 2.82 [2.78, 2.85] ▼ | 1.00 [1.00, 1.00] ▼ | 3.33 [3.33, 3.33] ▼ | 2.76 [2.73, 2.80] ▼ | 2.33 [2.33, 2.33] ▼ | 3.00 [3.00, 3.00] ▼ |
| | Intuitiveness | 3.71 [3.06, 4.37] ○ | 3.74 [3.72, 3.77] ▼ | 4.67 [4.67, 4.67] ▲ | 4.00 [4.00, 4.00] ○ | 3.81 [3.77, 3.84] ▼ | 3.07 [3.04, 3.09] ▼ | 3.00 [3.00, 3.00] ▼ |
| Contextual dimension | ***Average*** | 2.83 [2.12, 3.54] ▼ | 3.55 [3.53, 3.57] ▼ | 3.56 [3.54, 3.57] ▼ | 3.00 [3.00, 3.00] ▼ | 2.67 [2.65, 2.68] ▼ | 2.33 [2.33, 2.33] ▼ | 1.87 [1.87, 1.87] ▼ |
| | Name brand | 1.70 [1.31, 2.09] ▼ | 1.87 [1.84, 1.91] ▼ | 2.00 [2.00, 2.00] ▼ | 1.67 [1.67, 1.67] ▼ | 2.00 [2.00, 2.00] ▼ | 1.67 [1.67, 1.67] ▼ | 1.00 [1.00, 1.00] ▼ |
| | Mass market | 2.73 [1.65, 3.81] ▼ | 2.65 [2.63, 2.66] ▼ | 4.33 [4.33, 4.33] ▲ | 3.33 [3.33, 3.33] ▼ | 2.07 [2.04, 2.09] ▼ | 2.67 [2.67, 2.67] ▼ | 1.33 [1.33, 1.33] ▼ |
| | Fashionability | 2.61 [1.80, 3.42] ▼ | 4.05 [4.01, 4.09] ▲ | 2.45 [2.38, 2.51] ▼ | 2.00 [2.00, 2.00] ▼ | 2.81 [2.76, 2.87] ▼ | 2.33 [2.33, 2.33] ▼ | 2.00 [2.00, 2.00] ▼ |
| | Social approval | 2.86 [1.99, 3.73] ▼ | 3.59 [3.53, 3.66] ▼ | 4.00 [4.00, 4.00] ○ | 3.00 [3.00, 3.00] ▼ | 2.55 [2.51, 2.60] ▼ | 2.00 [2.00, 2.00] ▼ | 2.00 [2.00, 2.00] ▼ |
| | Credibility | 4.24 [3.09, 5.40] ○ | 5.56 [5.53, 5.60] ▲ | 5.00 [5.00, 5.00] ▲ | 5.00 [5.00, 5.00] ▲ | 3.90 [3.87, 3.94] ▼ | 3.00 [3.00, 3.00] ▼ | 3.00 [3.00, 3.00] ▼ |

## B Rankings of LLM creativity evaluation standards by model

| Rank | LLM average | GPT | Grok | DeepSeek | ERNIE | Claude | Gemini |
|---|---|---|---|---|---|---|---|
| 1 | Paradigm shift 6.77 ▲ | Breakthrough 7.00 ▲ | Breakthrough 7.00 ▲ | Paradigm shift 7.00 ▲ | Breakthrough 6.57 ▲ | Paradigm shift 6.33 ▲ | Breakthrough 7.00 ▲ |
| 2 | Potential 6.62 ▲ | Paradigm shift 7.00 ▲ | Paradigm shift 7.00 ▲ | Potential 6.67 ▲ | Paradigm shift 6.28 ▲ | Potential 6.00 ▲ | Paradigm shift 7.00 ▲ |
| 3 | Breakthrough 6.59 ▲ | Potential 7.00 ▲ | Potential 7.00 ▲ | Breakthrough 6.33 ▲ | Rarity 6.15 ▲ | Repurposing 6.00 ▲ | Potential 7.00 ▲ |
| 4 | Repurposing 6.22 ▲ | Rarity 6.68 ▲ | Rarity 7.00 ▲ | Repurposing 5.67 ▲ | Potential 6.07 ▲ | Breakthrough 5.67 ▲ | Rarity 6.67 ▲ |
| 5 | Rarity 6.19 ▲ | Repurposing 6.67 ▲ | Repurposing 7.00 ▲ | Combination 5.00 ▲ | Combination 6.00 ▲ | Combination 5.67 ▲ | Combination 6.33 ▲ |
| 6 | Combination 6.00 ▲ | Combination 6.36 ▲ | Combination 6.66 ▲ | Credibility 5.00 ▲ | Repurposing 6.00 ▲ | Rarity 5.67 ▲ | Repurposing 6.00 ▲ |
| 7 | Updates tradition 5.54 ▲ | Updates tradition 6.33 ▲ | Artistry 6.27 ▲ | Experience 5.00 ▲ | Experience 5.67 ▲ | Artistry 5.33 ▲ | Updates tradition 5.67 ▲ |
| 8 | Artistry 5.42 ▲ | Artistry 6.27 ▲ | Updates tradition 6.00 ▲ | Rarity 5.00 ▲ | Artistry 5.66 ▲ | Updates tradition 5.33 ▲ | Experience 5.33 ▲ |
| 9 | Experience 5.39 ▲ | Joy 5.77 ▲ | Surprise 5.80 ▲ | Artistry 5.00 ▲ | Updates tradition 5.26 ▲ | Experience 5.00 ▲ | Social interaction 4.67 ▲ |
| 10 | Variety 5.03 ▲ | Experience 5.68 ▲ | Experience 5.67 ▲ | Functionality 4.67 ▲ | Variety 5.00 ▲ | Joy 5.00 ▲ | Variety 4.67 ▲ |
| 11 | Joy 4.84 ▲ | Functionality 5.67 ▲ | Functionality 5.67 ▲ | High tech 4.67 ▲ | Surprise 4.82 ▲ | Variety 4.69 ▲ | Joy 4.33 ▲ |
| 12 | Surprise 4.59 ○ | Credibility 5.56 ▲ | Variety 5.64 ▲ | Updates tradition 4.67 ▲ | Social interaction 4.65 ▲ | Surprise 4.67 ▲ | Artistry 4.00 ○ |
| 13 | Social interaction 4.55 ○ | Variety 5.51 ▲ | Joy 5.33 ▲ | Variety 4.67 ▲ | Joy 4.62 ▲ | Functionality 3.67 ▼ | Feasibility 4.00 ○ |
| 14 | Functionality 4.55 ○ | Social interaction 5.33 ▲ | Credibility 5.00 ▲ | Social interaction 4.33 ▲ | High tech 4.33 ▲ | Social interaction 3.67 ▼ | Ease of use 3.67 ▼ |
| 15 | Credibility 4.24 ○ | Surprise 5.27 ▲ | Wide use 4.67 ▲ | Ease of use 4.00 ○ | Ease of use 4.00 ○ | Wide use 3.67 ▼ | Functionality 3.67 ▼ |
| 16 | High tech 4.11 ○ | Ease of use 4.98 ▲ | Intuitiveness 4.67 ▲ | Intuitiveness 4.00 ○ | Functionality 3.95 ▼ | High tech 3.33 ▼ | Surprise 3.67 ▼ |
| 17 | Ease of use 4.03 ○ | High tech 4.98 ▲ | Social interaction 4.67 ▲ | Joy 4.00 ○ | Credibility 3.90 ▼ | Ease of use 3.25 ▼ | Credibility 3.00 ▼ |
| 18 | Intuitiveness 3.71 ○ | Harmony 4.50 ▲ | High tech 4.33 ▲ | Wide use 4.00 ○ | Intuitiveness 3.81 ▼ | Intuitiveness 3.07 ▼ | High tech 3.00 ▼ |
| 19 | Wide use 3.68 ○ | Wide use 4.30 ▲ | Mass market 4.33 ▲ | Feasibility 3.67 ▼ | Harmony 3.04 ▼ | Credibility 3.00 ▼ | Intuitiveness 3.00 ▼ |
| 20 | Harmony 3.37 ○ | Fashionability 4.05 ▲ | Ease of use 4.29 ▲ | Harmony 3.67 ▼ | Fashionability 2.81 ▼ | Feasibility 2.67 ▼ | Observability 3.00 ▼ |
| 21 | Feasibility 3.35 ▼ | Feasibility 3.78 ▼ | Harmony 4.00 ○ | Mass market 3.33 ▼ | Wide use 2.78 ▼ | Harmony 2.67 ▼ | Wide use 2.67 ▼ |
| 22 | Social approval 2.86 ▼ | Intuitiveness 3.74 ▼ | Social approval 4.00 ○ | Observability 3.33 ▼ | Observability 2.76 ▼ | Mass market 2.67 ▼ | Harmony 2.33 ▼ |
| 23 | Mass market 2.73 ▼ | Social approval 3.59 ▼ | Feasibility 3.42 ▼ | Surprise 3.33 ▼ | Social approval 2.55 ▼ | Fashionability 2.33 ▼ | Fashionability 2.00 ▼ |
| 24 | Fashionability 2.61 ▼ | Observability 2.82 ▼ | Fashionability 2.45 ▼ | Social approval 3.00 ▼ | Feasibility 2.54 ▼ | Observability 2.33 ▼ | Social approval 2.00 ▼ |
| 25 | Observability 2.54 ▼ | Mass market 2.65 ▼ | Name brand 2.00 ▼ | Fashionability 2.00 ▼ | Mass market 2.07 ▼ | Social approval 2.00 ▼ | Mass market 1.33 ▼ |
| 26 | Name brand 1.70 ▼ | Name brand 1.87 ▼ | Observability 1.00 ▼ | Name brand 1.67 ▼ | Name brand 2.00 ▼ | Name brand 1.67 ▼ | Name brand 1.00 ▼ |

**Importance rating**

1 4 7

**95% CI relative to scale midpoint (4)**

- ▲ Entirely above midpoint
- ○ Includes midpoint
- ▼ Entirely below midpoint

**Figure 3. Alignment between LLM and human creativity evaluations.** Each estimate represents the Pearson correlation between LLM creativity ratings and the final human creativity scores across 1,103 ideas. The LLM average estimate, shown as a diamond, represents the correlation between creativity ratings averaged across the six LLMs and the final human creativity scores. Model-specific estimates, shown as circles, represent the correlations between each individual LLM's creativity ratings and the final human creativity scores. Error bars indicate 95% confidence intervals.

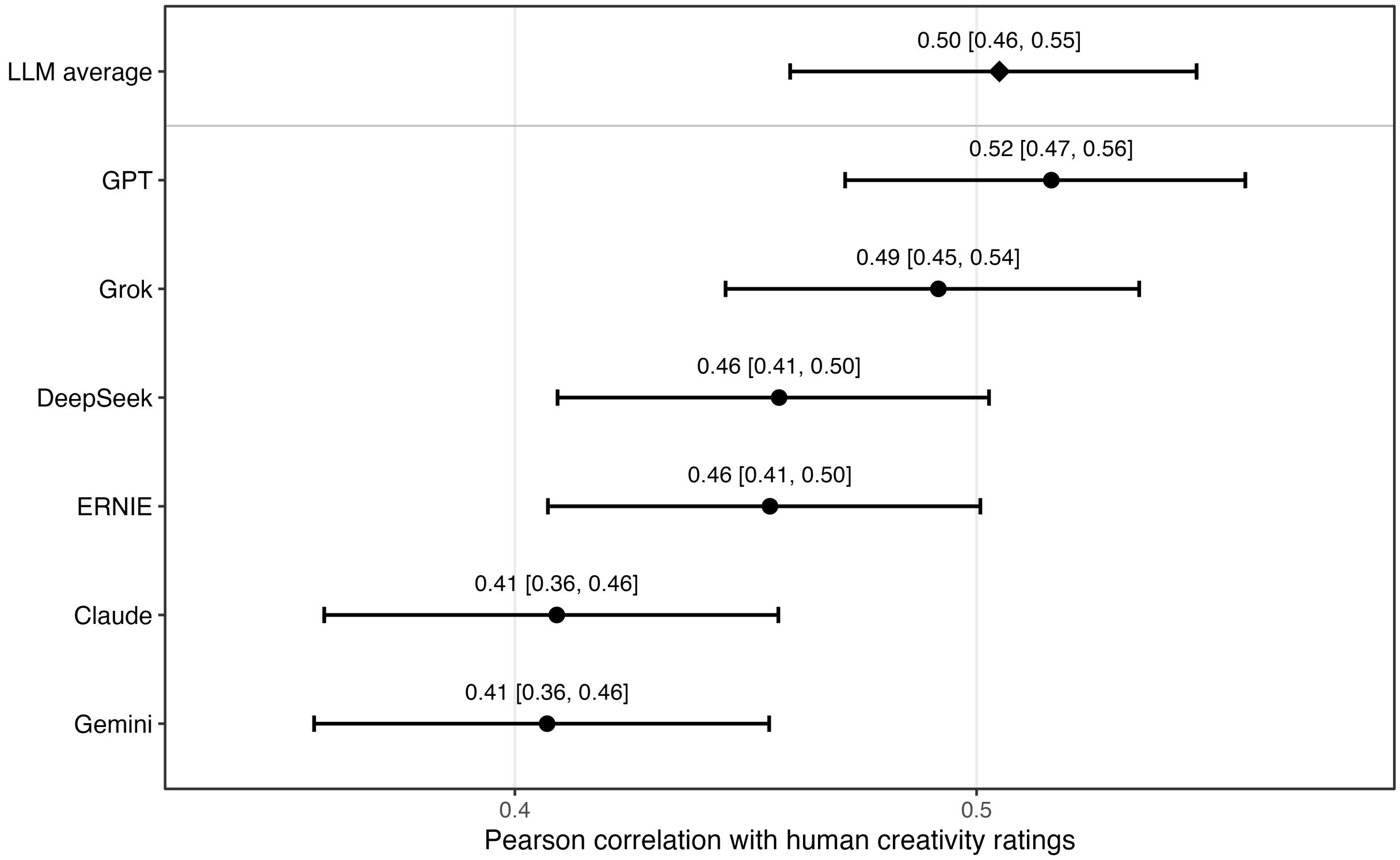

**Figure 4. Contextual information differentially influenced LLM and human creativity evaluations.** Estimated mean-centred creativity ratings are shown by Product Description and Evaluator Source. Error bars indicate 95% confidence intervals.

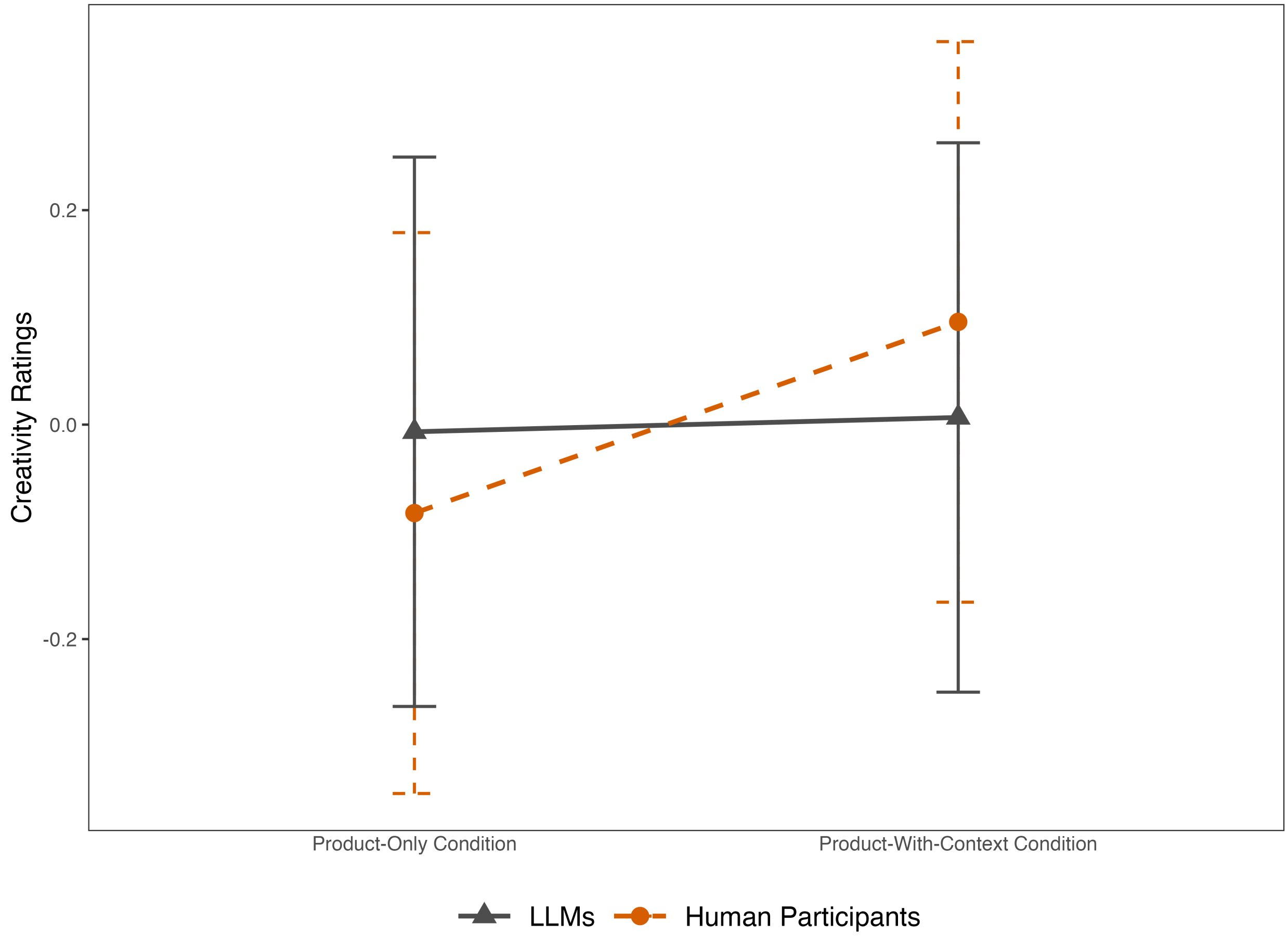

**Figure 5. Reduced contextual sensitivity was consistent across individual LLMs.** Each panel compares creativity ratings from one LLM with ratings from human participants. Estimated mean-centred creativity ratings are shown by Product Description and Evaluator Source. Error bars indicate 95% confidence intervals.

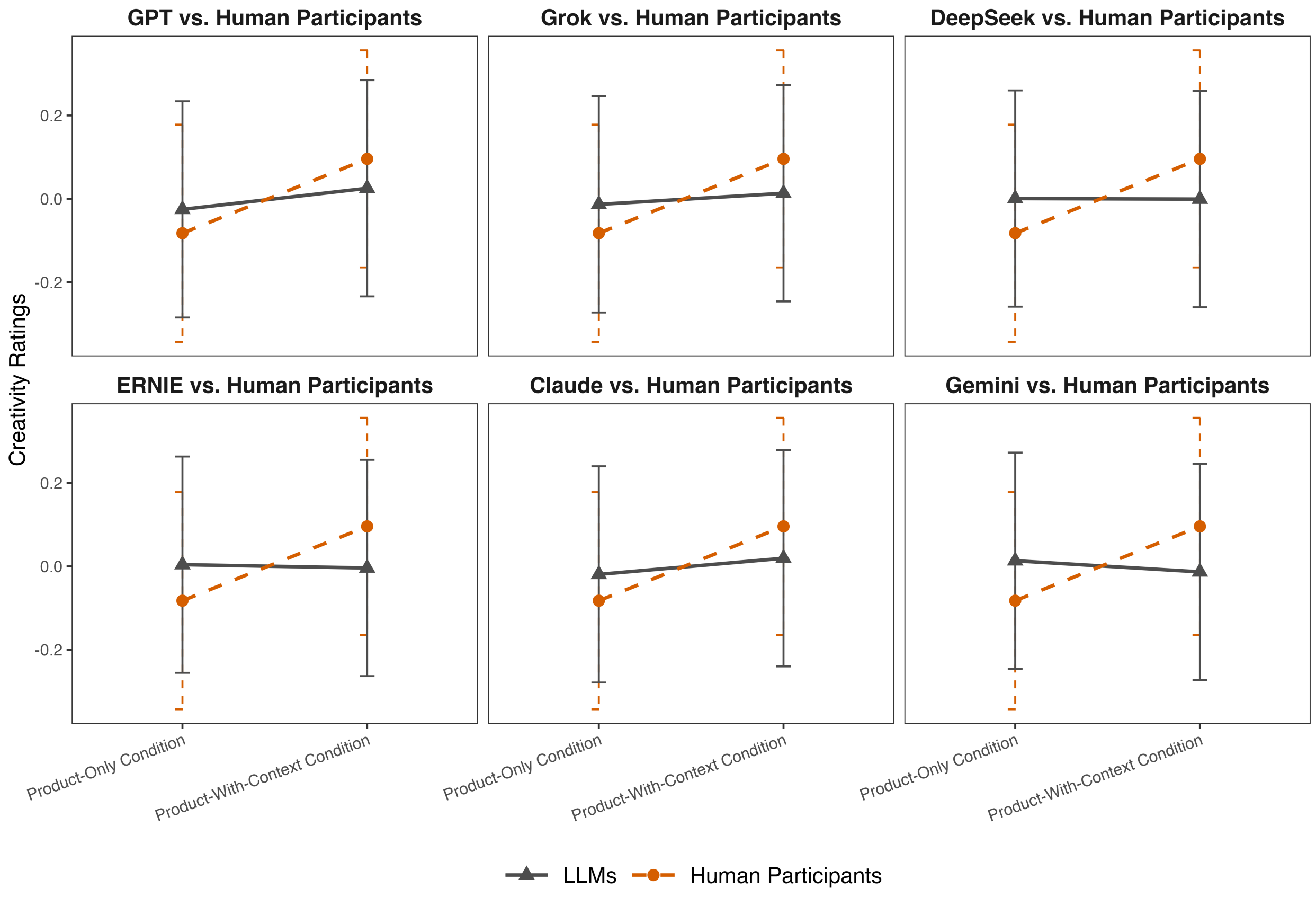

**Figure 6. Differences in contextual sensitivity between LLMs and human participants across contextual variants.** Each estimate represents the interaction coefficient comparing the effect of a contextual variant on creativity evaluations from LLMs and human participants, relative to the product-only condition. Error bars indicate 95% confidence intervals. Negative values indicate that the contextual variant produced a smaller change in LLM evaluations than in human evaluations. The dashed vertical line marks zero difference between LLMs and human participants.

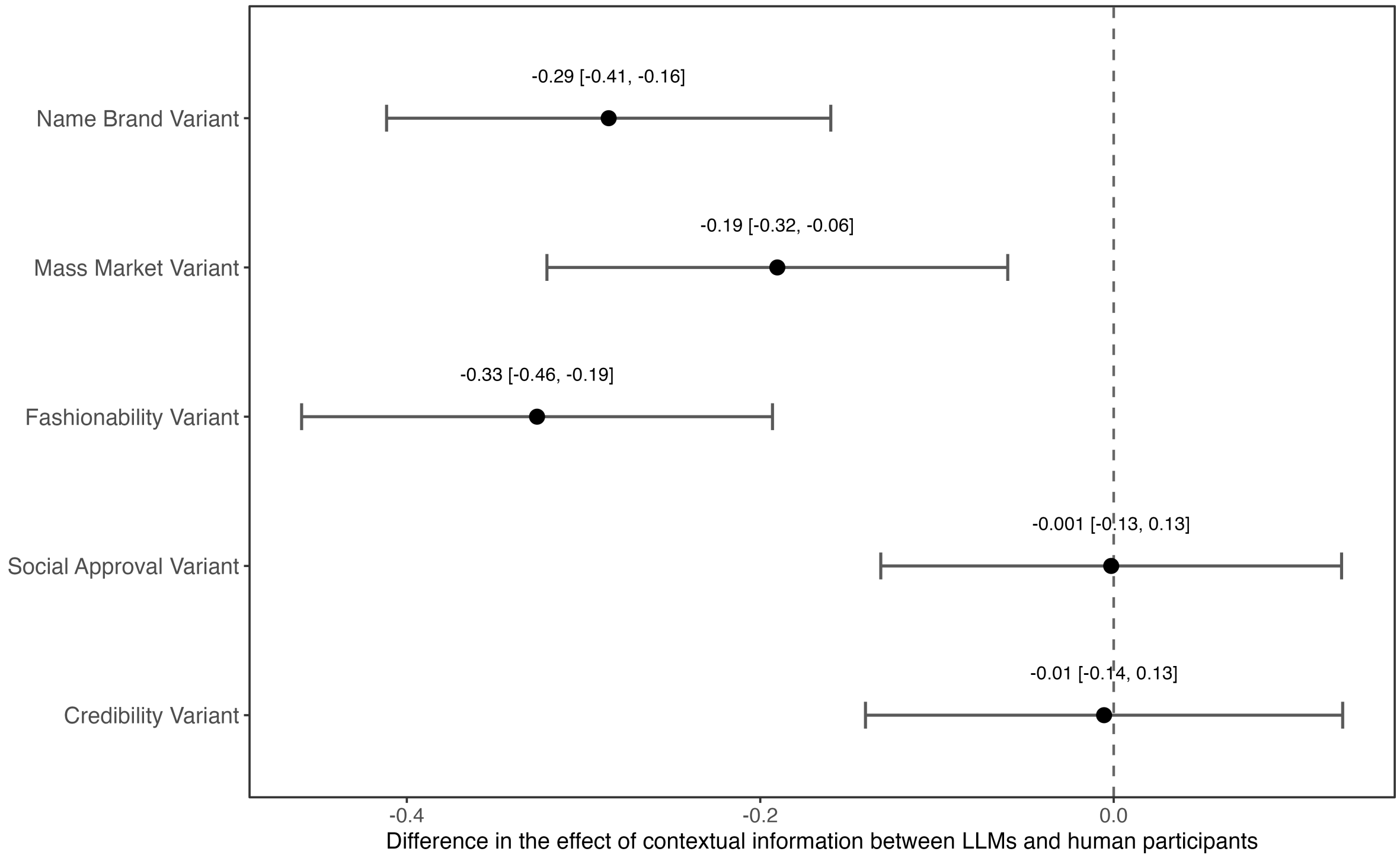

**Table 1. Dimensions of creativity evaluation standards.**

| Dimension | Description | Creativity Evaluation Standards |
|---|---|---|
| Novelty dimension | This dimension captures the extent to which an idea is original, unconventional, unexpected, or meaningfully different from existing alternatives. | Breakthrough; Combination; Paradigm shift; Rarity; Repurposing; Updates tradition; Potential; Surprise |
| Usefulness dimension | This dimension captures the extent to which an idea can serve a valued purpose, address a problem, or work appropriately in use. | Ease of use; Feasibility; Functionality; Variety; Wide use |
| Hedonic dimension | This dimension captures the extent to which an idea creates positive experiential, aesthetic, or affective value. | Artistry; Joy; Experience |
| Relational dimension | This dimension captures the extent to which an idea creates, supports, or harmonises relationships among people. | Social interaction; Harmony |
| Technology dimension | This dimension captures the extent to which an idea draws on advanced, cutting-edge, or technologically sophisticated capabilities. | High tech |
| Concreteness dimension | This dimension captures the extent to which an idea is specific, tangible, observable, or easy to mentally represent. | Observability; Intuitiveness |
| Contextual dimension | This dimension captures the broader social, market, or reputational environment surrounding an idea. | Name brand; Mass market; Fashionability; Social approval; Credibility |

## Methods

All studies were approved by the Ethics Review Board of the first author's institution. All human participants provided informed consent. We collected all LLM responses via each provider's API and conducted the cultural consensus theory analysis in Python (version 3.12.7). All other statistical analyses were conducted in R (version 4.5.3).

### *Study 1*

Study 1 evaluated the creativity evaluation standards of six state-of-the-art LLMs: GPT-4.1 mini[21], Grok 4 Fast[22], DeepSeek V3[23], ERNIE 3.5-8K[24], Claude Haiku 4.5[25], and Gemini 2.5 Flash-Lite[26]. To assess these standards, we used an established measure of human creativity evaluation that identifies 26 standards people use when judging creativity[17], with each standard capturing a type of information that can function as evidence of creativity (see Supplementary Table 1 for standards and their definitions). Each LLM was prompted with the question, "How important is this feature to a product being creative?" to assess the importance of each standard on a 7-point Likert scale (1 = not important at all, 7 = extremely important). For example, a sample item measuring the importance of the "combination" standard was "It integrates opposing functions or features" (see Supplementary Information section B for all items).

Consistent with prior research[37], we set the temperature parameter to 0 to improve reproducibility and reduce stochastic variation in model responses (this setting reduces but does not eliminate response variability across identical API calls). To account for the residual variability, each model completed 100 independent iterations for each item, yielding 100 ratings per item. Each item was presented separately using the same prompt structure and model settings across all LLMs (see Supplementary Information section C for the LLM prompts used in Study 1). Each rating was collected through a separate API call and conversation context so that a model's response to one item could not influence its response to another item. For each model, we first averaged ratings across the 100 iterations for each item and then averaged item ratings within each standard to obtain the final importance score for that standard.

***Study 2***

Study 2 examined the downstream consequences of standard convergence and proceeded in two parts: idea generation (Part 1) and idea evaluation (Part 2). In Part 1, we recruited 293 human participants (107 males, 153 females, 7 non-binary/third gender, and 26 participants chose not to disclose their gender; $M_{age}$ = 28.26, $SD_{age}$ = 10.35) from a leading UK university participant pool to complete three creativity tasks. Participants were compensated at a rate of £10 per hour. The tasks included two in-basket exercises featuring real-world business scenarios[32] and one environmental task related to water conservation[33]. Participants generated one idea for each in-basket task and three ideas for the environmental task. After excluding blank and nonsensical responses (e.g., gibberish text), we retained 1,103 valid ideas for LLM and human evaluation.

In Part 2, we obtained creativity evaluations for the 1,103 ideas from human evaluators and LLMs. Following the consensual assessment technique[15], we recruited three trained human evaluators to assess the creativity of each idea. Each evaluator independently rated the extent to which each idea was creative on a 7-point Likert scale (1 = completely not creative, 7 = completely creative). The three evaluators showed substantial agreement in their creativity ratings (ICC1 = 0.45, ICC2 = 0.71, median rWG = 0.89)[44,45]. We therefore averaged their ratings to obtain the final human creativity score for each idea.

The six LLMs rated the same set of ideas using the same 7-point Likert scale as the human evaluators (see Supplementary Information section F for the LLM prompts used in Study 2). To improve the stability of these ratings, each model completed five independent evaluation iterations for each task. In each iteration, the ideas were shuffled into a random order and split into batches of 25, with the final batch containing fewer ideas when necessary. The LLM then evaluated each batch separately. Because the order and grouping of ideas changed with each iteration, it helped ensure that an idea's rating was not skewed by which other ideas happened to appear alongside it or by its position in the sequence. Finally, for each model, we averaged the five ratings for each idea to obtain the model's final creativity score for that idea.

***Study 3***

Study 3 (preregistered at https://aspredicted.org/md2m2j.pdf) examined the downstream consequences of standard divergence. We recruited 1,500 participants from Prolific and excluded 305 based on a preregistered attention check that asked them to identify the information provided in the product description, leaving a final sample of 1,195 participants (658 males, 530 females, 5 non-binary/third gender, and 2 participants chose not to disclose their gender; $M_{\text{age}} = 34.54$, $SD_{\text{age}} = 12.14$). Participants were compensated at a rate of £6 per hour.

Study 3 employed a 2 (Product Description: product-only vs. product-with-context) × 2 (Evaluator Source: human vs. LLM) design. The Product Description factor varied whether evaluators received only information about the product itself or the same information supplemented with contextual information about the product. Specifically, in the product-only condition, evaluators received a base product description that described the product's features, functions, components, and use. In the product-with-context condition, evaluators received the same base product description plus one additional sentence that provided contextual information about the product. The Evaluator Source factor varied whether the creativity evaluation was provided by human participants or by LLMs.

We constructed the product descriptions in two steps. First, we identified 15 products from Kickstarter, a crowdfunding platform widely used for creative products. These products spanned multiple categories such as household appliances, furniture, and consumer electronics, and had not yet become available to consumers at the time of data collection, thereby reducing the likelihood that participants had direct prior experience with them. For each product, we created a base description using materials provided by the product's creator team. Each base description focused on the intrinsic characteristics of the product, such as functionality, components, and use. These base descriptions were used in the product-only condition. Second, we created five one-sentence contextual variants, each corresponding to one of the five contextual standards examined in Study 1. Each contextual variant provided one type of contextual information about the product, such as its market positioning, brand prestige, fashionability, social approval, or source credibility. In the product-with-context condition, evaluators saw the same base description with one randomly selected contextual variant appended. Thus, for each product, we prepared one base description and five additional descriptions that paired the base description with a different contextual variant. Full product descriptions and contextual variants are reported in Supplementary Tables 3 and 4.

Human participants were randomly assigned to one of the 15 products and to either the product-only condition or the product-with-context condition. Participants in the product-only condition read the base description of the assigned product, whereas those in the product-with-context condition read the same base description plus one randomly selected contextual variant. After reading the product description, participants completed an attention check question and rated the product's creativity on a 7-point scale (1 = not at all creative, 7 = extremely creative). The attention check asked participants to identify which piece of information had appeared in the product description. Participants selected from seven options: five listed the contextual variants, one stated “none of the above,” and one stated, “I do not remember.” Participants in the product-with-context condition failed the attention check if they did not select the contextual variant they had seen. Participants in the product-only condition failed the attention check if they selected one of the contextual variants.

For the LLM evaluations, each model evaluated the same set of product descriptions used in the human evaluation task (see Supplementary Information section I for the LLM prompts used in Study 3). For each product, this set included the base description and five additional descriptions, each consisting of the same base description paired with one contextual variant. Each LLM provided creativity ratings using the same scale as human participants. To align with the sampling structure of the human evaluation, each LLM rated the base description 50 times for each product in the product-only condition. In the product-with-context condition, each LLM rated each of the five additional versions 10 times, again yielding 50 ratings per product. Each evaluation was conducted in a separate prompt and conversation context to avoid carryover across conditions.

**Data Availability.** The preregistrations, survey instruments and data for all studies are available at the Open Science Framework:

https://osf.io/q4xa9/overview?view_only=0e22435638ce41468c1f8303f4bab0a9.

**Code Availability.** The code, written in R and Python for analyses and data visualisations, is available at the Open Science Framework:

https://osf.io/q4xa9/overview?view_only=0e22435638ce41468c1f8303f4bab0a9.

**References**

44. LeBreton, J. M. & Senter, J. L. Answers to 20 Questions About Interrater Reliability and Interrater Agreement. *Organ. Res. Methods* **11**, 815–852 (2008).

45. Cicchetti, D. V. Guidelines, criteria, and rules of thumb for evaluating normed and standardized assessment instruments in psychology. *Psychol. Assess.* **6**, 284–290 (1994).

**Competing Interests.** The authors declare no competing interests.

**Supplementary Information.** Supplementary Information is available for this paper.